\documentclass[a4paper, 10pt, conference]
{ieeeconf}

\IEEEoverridecommandlockouts
\usepackage{amsmath}
\usepackage{amssymb}
\usepackage{pgfplots}
\usepackage{subcaption}
\usepackage{soul}
\usepackage{graphicx}

\title{\LARGE \bf
Data-Driven Dynamic Parameter Learning of manipulator robots
} 

\author{
Mohammed Elseiagy$^{1,3}$, Tsige Tadesse Alemayoh$^{1,2}$, Ranulfo Bezerra$^{1,2}$, Shotaro Kojima$^{1,2}$, Kazunori Ohno$^{1,2}$%
\thanks{$^{1}$ Graduate School of Information Sciences, Tohoku University, Sendai, Japan}%
\thanks{$^{2}$ Tough Cyberphysical AI Research Center, Tohoku University, Sendai, Japan}%
\thanks{$^{3}$ Egypt-Japan University of Science and Technology (E-JUST), New Borg El-Arab, Alexandria, Egypt}%
\thanks{{\tt mohammed.abelaziz@ejust.edu.eg }}%
}

\begin{document}

\maketitle
\thispagestyle{empty}
\pagestyle{empty}


\begin{abstract} 
Bridging the sim-to-real gap remains a fundamental challenge in robotics, as accurate dynamic parameter estimation is essential for reliable model-based control, realistic simulation, and safe deployment of manipulators. Traditional analytical approaches often fall short when faced with complex robot structures and interactions. Data-driven methods offer a promising alternative, yet conventional neural networks such as recurrent models struggle to capture long-range dependencies critical for accurate estimation. In this study, we propose a Transformer-based approach for dynamic parameter estimation, supported by an automated pipeline that generates diverse robot models and enriched trajectory data using Jacobian-derived features. The dataset consists of 8,192 robots with varied inertial and frictional properties. Leveraging attention mechanisms, our model effectively captures both temporal and spatial dependencies. Experimental results highlight the influence of sequence length, sampling rate, and architecture, with the best configuration (sequence length 64, 64 Hz, four layers, 32 heads) achieving a validation $R^{2}$ of 0.8633. Mass and inertia are estimated with near-perfect accuracy, Coulomb friction with moderate-to-high accuracy, while viscous friction and distal link center-of-mass remain more challenging. These results demonstrate that combining Transformers with automated dataset generation and kinematic enrichment enables scalable, accurate dynamic parameter estimation, contributing to improved sim-to-real transfer in robotic systems
\end{abstract}

\section{Introduction}

Robotic arms are essential in today’s industries and research, carrying out tasks such as precision manufacturing, surgical support, and exploration of dangerous environments \cite{wu2022review}. The performance and safety of these systems depend directly on the accuracy of their dynamic models. These models, defined by parameters such as mass, inertia, friction, and damping, determine how the robot moves and interacts with its surroundings. Estimating these dynamic parameters correctly is not just a theoretical exercise; it is a key requirement for enabling advanced robotic functions and solving important real-world problems \cite{lee2024robot}.

One of the main challenges in robotics is the so-called "reality gap" in \textbf{sim-to-real transfer}. Training control policies or reinforcement learning agents directly on physical robots is often difficult due to safety risks, limited time, and hardware costs. Simulations provide a safe and efficient alternative, but controllers trained in simulation often fail in the real world because of differences between simulated and real dynamics \cite{rothert2024sim, zhao2020sim, lu2022pose}. Accurate dynamic parameter estimation is crucial to close this gap. By carefully identifying a robot’s physical properties, we can build simulations that closely match real physics, making sim-to-real transfer more reliable and enabling advanced control and learning methods on real systems \cite{bargellini_sim2real_2023, ren2023adaptsim, garg2024dynamics}. In addition, accurate models are the basis of \textbf{digital twins}, which are real-time virtual copies of physical systems used for monitoring, predictive maintenance, and optimization \cite{chinnasamy2023digital, liu2024hybrid, chen2023designing, wang2024research}. For robotic arms, a high-quality digital twin allows safe testing of new control algorithms, early detection of wear and performance loss, and even remote fault diagnosis, improving efficiency and extending system life.

Despite its importance, estimating these parameters remains challenging. Traditional analytical methods, such as least-squares estimation, often struggle with the complex and coupled dynamics of multi-joint arms \cite{wu2022review}. These methods are sensitive to noise and can produce results that are not physically meaningful \cite{lee2024robot}. Moreover, dynamic parameters can change over time because of wear, temperature shifts, or changes in payload, requiring frequent re-calibration. Data collection in real robots adds further difficulty due to sensor noise and the challenge of obtaining accurate measurements.

To overcome these issues, data-driven methods, especially those based on deep learning, have become strong alternatives. These approaches can learn system dynamics directly from sensor data without requiring explicit physical models. While Recurrent Neural Networks (RNNs) and Long Short-Term Memory (LSTM) networks have shown success in capturing time-based patterns \cite{LSTMs, polydoros2015real, segota2022dynamics}, they often struggle with very long dependencies and may generalize poorly. The \textbf{Transformer} architecture, originally developed for natural language processing \cite{vaswani2017attentionneed}, offers a promising solution. Its self-attention mechanism is effective at capturing long-range dependencies in sequence data, making it suitable for the complex dynamics of robotic arms \cite{ReedTransformer, shukla2021multitime, Foumani_2023}. By focusing attention on key joint interactions and using cross-attention across joints, Transformers can build strong and generalizable models of robot dynamics.

To address these challenges and push forward dynamic parameter estimation, we propose a complete pipeline that combines automated data generation, data enrichment, and Transformer-based model training. We first generate robot URDFs with varied dynamic properties using an automatic script. These URDFs are then simulated in Gazebo to create trajectory data through PID-controlled motion. The raw data is post-processed to add useful features such as kinematic information from Jacobian matrices, enriching the dataset. Finally, this dataset is used to train a new Transformer-based model built specifically for parameter estimation. By simulating a wide range of conditions—including changes in friction, damping, and inertia—our method aims to generalize well to real robots and allow rapid re-calibration when parameters change. This is highly relevant for applications such as collaborative robotics, surgical automation, and advanced manufacturing, where accurate models are vital for safety, efficiency, and adaptability.

The main contributions of this work are:
\begin{itemize}
  \item An automated pipeline for generating diverse robot models and trajectory data in simulation, incorporating gravity-aware PID control to produce rich datasets for parameter estimation.
  \item A Transformer-based architecture for accurate and generalizable dynamic parameter estimation, leveraging attention mechanisms to capture complex temporal and spatial dependencies in robot dynamics.
\end{itemize}

\section{Related Work}

The problem of estimating robot dynamic parameters has been studied from classical mechanics to modern deep learning. This section places our work in this context, showing the progress of approaches and clarifying our unique contribution.

\subsection{Classical and Analytical Identification Methods}
Traditional approaches are based on rigid body dynamics, often expressed with Euler-Lagrange equations. Weighted least-squares (WLS) methods are common, where carefully designed excitation trajectories are used to estimate base parameters \cite{wu2022review}. While effective for simple systems, these methods face serious challenges. They require special trajectories that can be unsafe for real robots, they cannot model complex non-linear effects such as friction well, and they are highly sensitive to noise, often producing unrealistic parameter values \cite{lee2024robot}. Grey-box methods, which combine physical models with data-driven techniques, have recently been proposed to improve robustness \cite{li2024evaluating}, especially for digital twins, but they still rely on simplified friction models or large experimental setups.

\subsection{Deep Learning for Robot Dynamics}
With deep learning, data-driven methods have become strong alternatives. They can learn dynamics directly from sensor data, removing the need for explicit physical equations. Early work used feed-forward networks, but the sequential nature of motion data led to RNNs and LSTMs. For example, \cite{LSTMs} and \cite{polydoros2015real} used LSTMs to estimate dynamic parameters from torque signals, showing they could capture time-based patterns. \cite{segota2022dynamics} studied machine learning methods based on synthetic data for dynamics modeling. However, recurrent models are hard to train, suffer from gradient issues, and often fail with very long sequences, limiting their ability to generalize across different robots and conditions.

\subsection{Transformers in Robotics and Time-Series Analysis}
The Transformer \cite{vaswani2017attentionneed} has recently proven powerful for sequence modeling beyond language, due to its self-attention mechanism. This mechanism allows it to consider all past states when predicting the next, overcoming the limitations of RNNs. In robotics, Transformers have been used for state prediction and dynamics modeling \cite{ReedTransformer}. Recent research has adapted Transformers for multivariate time-series, with new position encoding methods especially useful for robotic data \cite{Foumani_2023, shukla2021multitime}. More broadly, meta-learning with neural networks has been explored to quickly adapt models to new robots with little data, aligning with our goal of generalizable parameter estimation \cite{bianchi2024robomorph}. Our work differs by focusing specifically on estimating physical parameters and by introducing an automated data generation and enrichment pipeline. This places our research at the intersection of deep learning and robot system identification, aiming for a scalable and reliable solution for tasks like sim-to-real transfer and adaptive control.

\vspace{-4pt}
\section{Methodology}
\begin{figure}[h]
\centering
\includegraphics[trim=18 12 18 12, clip, width=1\columnwidth]
{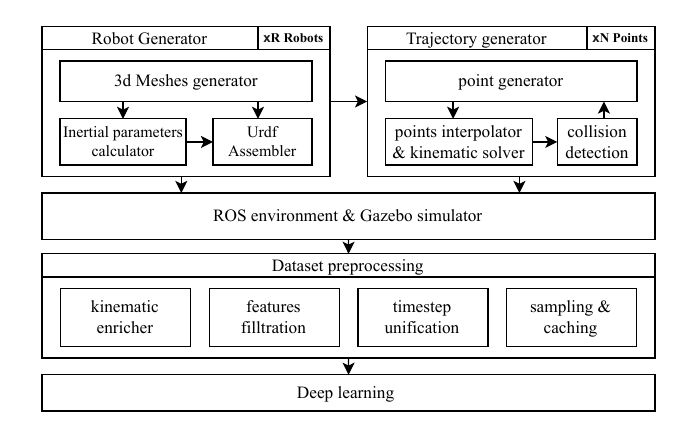}
\caption{The multi-stage pipeline for robotic data generation, simulation, preprocessing, and deep learning analysis.}
\label{fig:pipeline}
\end{figure}

\vspace{-2pt}
This research uses a multi-stage pipeline (Figure~\ref{fig:pipeline}) to generate, simulate, preprocess, and analyze robotic data with deep learning. The pipeline starts with the \textbf{Robot Generator}, which creates a diverse set of \emph{R Robots} by generating 3D meshes, computing inertial parameters, and assembling URDF models. In parallel, the \textbf{Trajectory Generator} produces \emph{N waypoints} per robot. Each waypoint is validated to avoid collisions by solving the forward kinematics, and interpolation ensures that intermediate points are also collision-free. These robots and trajectories are then integrated into a \textbf{ROS–Gazebo simulation} for realistic physical data. The simulated data undergoes \textbf{Dataset Preprocessing}, including kinematic enrichment, feature filtering, timestep unification, and sampling with caching. The final preprocessed dataset is then used for \textbf{Deep Learning} to extract insights and train predictive models.

\vspace{-24pt} 
\subsection{URDF Generator}
\begin{figure}[h]
\centering
\includegraphics[width=0.85\columnwidth]{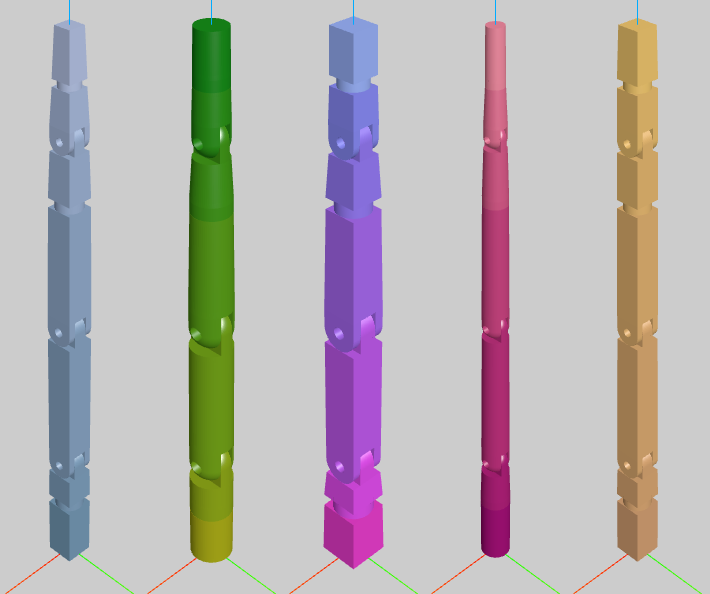}
\caption{Examples of robots generated by the URDF script. All robots share the same kinematic configuration (link lengths, joint axes, and offsets) but differ in their inertial and frictional dynamics due to changes in link cross-section, diameter, center of mass, and joint friction coefficients.}
\label{fig:robot}
\end{figure}

To construct the dataset, a custom URDF generator script was developed to produce large collections of robots. For each robot instance, the \textbf{kinematic configuration} (including link lengths, joint axes, and joint offsets) was held constant. However, the following structural and dynamic properties were systematically varied:
\begin{itemize}
  \item \textbf{Link cross-section shape}
  \item \textbf{Link diameter}
  \item \textbf{Link center of mass position}
  \item \textbf{Joint Coulomb friction coefficients}
  \item \textbf{Joint viscous friction coefficients}
\end{itemize}
These variations directly influence the robots’ inertial and frictional dynamics while preserving kinematic similarity. A total of \textbf{8,192 robots} were generated using this approach, examples of which are shown in Figure~\ref{fig:robot}.

\vspace{-14pt}
\subsection{Dynamic Parameters}
The dynamic parameters considered in this work are:
\begin{itemize}
  \item \textbf{Coulomb Friction}: This force opposes motion and is modeled as:
  \[ F_{\text{Coulomb}} = \mu_c \cdot \text{sign}(\dot{q}), \]
   where \(\mu_c\) is the Coulomb coefficient and \(\dot{q}\) is the joint velocity.
  \item \textbf{Viscous Friction}: This resistance is proportional to velocity and is modeled as:
  \[ F_{\text{Viscous}} = \mu_v \cdot \dot{q}, \]
  where \(\mu_v\) is the viscous coefficient.
  \item \textbf{Inertia Matrix}: This represents the mass distribution of each link and its resistance to acceleration.
\end{itemize}
These parameters were selected because they are challenging to measure directly and are susceptible to variations caused by environmental factors (e.g., temperature, humidity) and mechanical wear. In contrast, parameters such as link mass are relatively straightforward to measure and remain stable. The manipulator’s dynamics, including frictional effects, can be expressed by the following equation:
\[ M(q) \ddot{q} + C(q, \dot{q}) \dot{q} + G(q) + F_{\text{friction}} = \tau, \]
where \(M(q)\) is the inertia matrix, \(C(q,\dot{q})\) represents the Coriolis and centrifugal forces, \(G(q)\) is the gravity vector, \(\tau\) denotes the applied torques, and the total friction force \(F_{\text{friction}}\) is defined as:
\[ F_{\text{friction}} = \mu_c \cdot \text{sign}(\dot{q}) + \mu_v \cdot \dot{q}. \]

\subsection{Trajectory Generation}
For each robot, \textbf{16 workspace waypoints} were randomly selected. A per-joint PID controller was then utilized to guide the manipulator toward these waypoints. The controller was specifically designed and calibrated to achieve smooth and rapid transitions while preventing overshoot.

The simulator was configured to log raw data (joint positions, velocities, and applied torques) at a frequency of \textbf{1000 Hz}. To enhance robustness, the script incorporated anti-failure detection mechanisms to identify and automatically discard corrupted data resulting from collisions, simulator instabilities, or timeouts.

A standard PID controller without gravity compensation was initially used. 
However, to mitigate gravitational effects and improve control smoothness, 
a novel \textbf{gravity-aware PID modification} was introduced:  

\begin{equation}
u(t) = \left(K_p + K_G \frac{dj_i}{dz}\right) e(t) 
  + K_i \int_{0}^{t} e(\tau) \, d\tau 
  + K_d \frac{d e(t)}{dt},
\end{equation}

where \(K_G\) is a constant tuned to help the joint overcome gravity, 
and \(\tfrac{dj_i}{dz}\) is an element derived from the Jacobian matrix 
that indicates the effect of the motion of the \(i\)-th joint in the 
\(z\)-direction (the direction of gravity).  

This modification introduces a proportional torque adjustment based on 
the gravitational influence of each joint, leveraging Jacobian elements. 
The resulting hybrid compensation significantly improved trajectory 
tracking stability while keeping the controller lightweight.

\subsection{Post-Processing and Dataset Composition}
\begin{figure}[h]
\centering
\includegraphics[width=1\columnwidth]{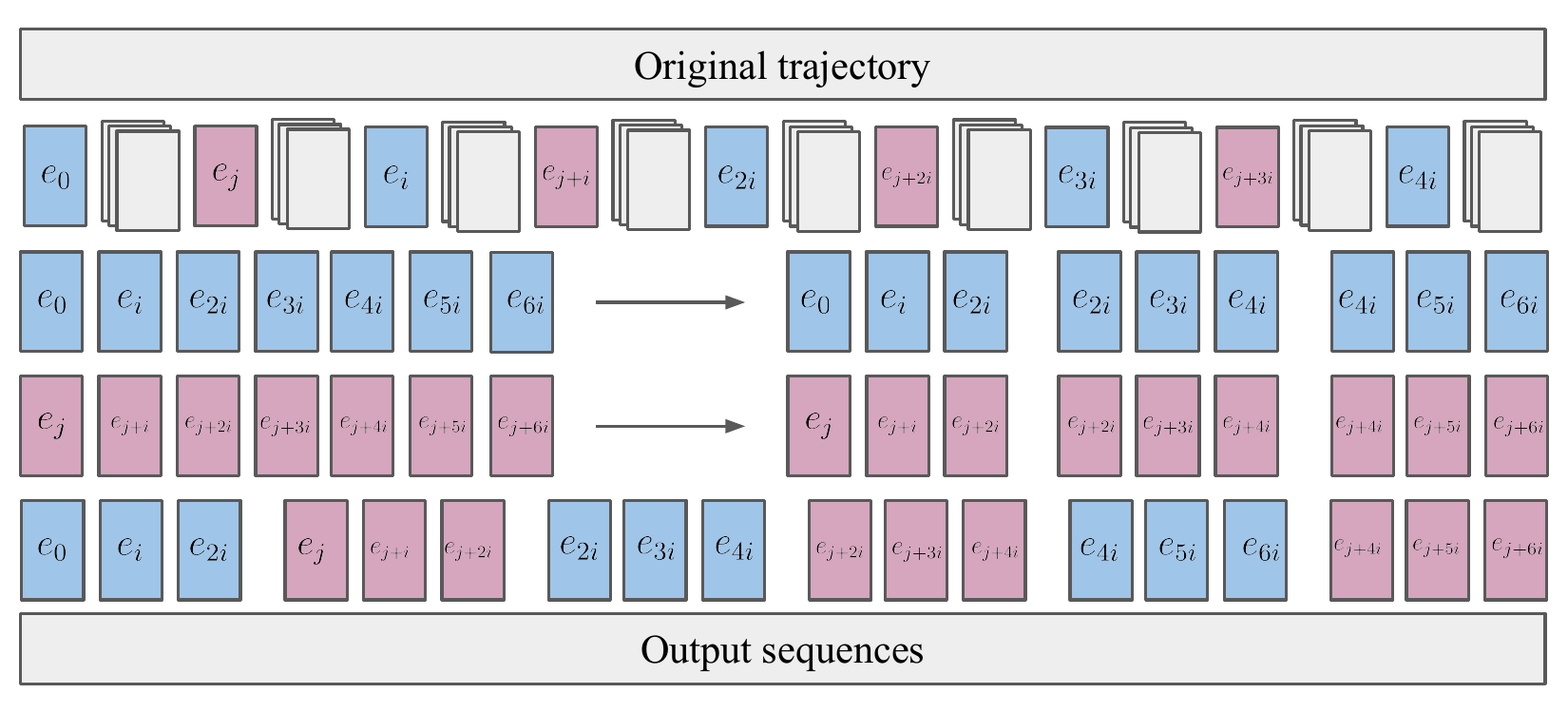}
\caption{Illustration of the offset-based sampling scheme. Multiple sequences are generated with temporal offsets and overlapping improving dataset utilization.}
\label{fig:sampling}
\end{figure}

Since the raw dataset was collected at 1000 Hz, it was too dense for direct use in model training. To make learning tractable, the trajectories were downsampled by selecting data at a smaller, fixed sampling frequency, thereby compressing the temporal information into manageable sequences. However, naive downsampling discards a significant portion of the collected dataset. To address this, a \textbf{secondary offset-based sampler} was implemented. This method extracts multiple sequences at the same sampling rate but with different temporal offsets, effectively reusing the raw data more efficiently. An illustration of this process is shown in Figure~\ref{fig:sampling}.

Additionally, the dataset was augmented with kinematic features:
\begin{itemize}
  \item \textbf{Jacobian Matrix Elements}: capturing the influence of joint motion on link positions.
  \item \textbf{Pruned Redundancies}: removal of constant or irrelevant terms (e.g., translational components for a base link that only rotates).
\end{itemize}

\vspace{-16pt}
\subsection{Final Dataset Composition}
The final dataset comprises:
\begin{itemize}
  \item \textbf{8,192 robots} with varied inertial and frictional properties.
  \item \textbf{16 trajectories per robot}, defined by randomly chosen workspace waypoints.
  \item Augmented kinematic features (filtered Jacobian elements per link).
\end{itemize}
This methodology ensures that the dataset captures a wide spectrum of dynamic behaviors while maintaining efficiency and robustness for learning-based parameter estimation.

\subsection{Transformer-Based Model Architecture}
The proposed model adopts the original transformer architecture introduced by Vaswani et al. \cite{vaswani2017attentionneed}, initially developed for natural language processing (NLP) and subsequently extended to sequential data tasks. Transformers are particularly effective for robotic time-series data due to their ability to capture long-range dependencies and temporal relationships, making them well-suited for dynamic parameter estimation. Our model consists of three key components: input embedding, joint-specific transformers, and output decoding. 

\begin{figure}[htbp]
  \centering
  \includegraphics[trim=0 30 0 40, clip, width=180pt]{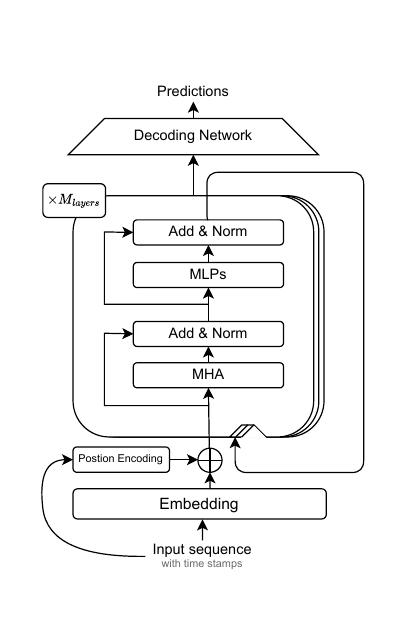}
  \caption{The architecture of the proposed model, based on the original transformer \cite{vaswani2017attentionneed}.}
  \label{fig:architecture}
\end{figure}

\subsubsection{Input Embedding}
Input features, including joint positions, velocities, torques, and kinematic elements, are embedded into a fixed-dimensional space for transformer processing. Adhering to the original transformer design \cite{vaswani2017attentionneed}, positional encoding is used to represent temporal information, rather than specialized encodings for irregular time series. To ensure compatibility, the raw dataset (originally sampled at 1000 Hz) was resampled onto a fixed frequency grid using interpolation, thereby enabling the use of standard positional encodings.

While alternative encoding schemes such as tAPE and eRPE \cite{Foumani_2023} have been proposed to better represent time in irregularly sampled sequences, and methods like \cite{shukla2021multitime} directly address irregular sampling using attention mechanisms, our focus was on maintaining fidelity to the original transformer formulation. Therefore, preprocessing steps (interpolation and resampling) were applied to regularize the data before input, preserving compatibility with the canonical architecture while still capturing essential temporal dynamics.

\section{Results}

This section presents a detailed evaluation of the proposed transformer-based approach for dynamic parameter estimation in robotic manipulators. We focus on three aspects: (i) the effect of dataset configuration, (ii) the impact of transformer architecture choices, and (iii) the performance of the best model on dynamic parameters.

\subsection{Effect of Dataset Configuration}

To assess how sequence length, sampling rate, and secondary sampling rate affect model performance, we trained models with the same architecture on different dataset configurations. Table~\ref{tab:dataset_ablation} summarizes the results, including validation $R^2$, effective input time per sequence, and dataset utilization:

\[
\text{Effective Time (s)} = \text{seq\_len} \times \frac{\text{sampling rate (Hz)}}{1000}
\]

\begin{table}[h!]
\centering
\caption{Experimental study on dataset configurations using fixed transformer architecture.}
\label{tab:dataset_ablation}
\begin{tabular}{|c|c|c|c|c|}
\hline
\textbf{Seq Len} & \textbf{Sampling Rate} & \textbf{SSR} & \textbf{Effective Time (s)} & \textbf{Val $R^2$} \\ \hline
16 & 32 & 8 & 0.512 & 0.7880 \\
16 & 64 & 16 & 1.024 & 0.8090 \\
16 & 128 & 32 & 2.048 & 0.8333 \\
16 & 256 & 32 & 4.096 & 0.8446 \\
32 & 32 & 8 & 1.024 & 0.8134 \\
32 & 64 & 16 & 2.048 & 0.8428 \\
32 & 128 & 128 & 16.384 & 0.8383 \\
64 & 32 & 8 & 2.048 & 0.8328 \\
64 & 32 & 16 & 2.048 & 0.8331 \\
64 & 64 & 16 & 4.096 & \textbf{0.8633} \\
64 & 64 & 32 & 4.096 & 0.8592 \\
64 & 64 & 64 & 4.096 & 0.8571 \\
128 & 16 & 8 & 2.048 & 0.8181 \\
128 & 32 & 8 & 4.096 & 0.8451 \\
128 & 32 & 16 & 4.096 & 0.8459 \\
128 & 64 & 16 & 8.192 & 0.8479 \\
\hline
\end{tabular}
\end{table}

The results show that increasing sequence length generally improves performance up to 64 steps, after which gains diminish.

\subsection{Transformer Architecture Comparison}

We investigated the impact of architectural choices using a fixed dataset configuration. Table~\ref{tab:arch_ablation} compares models with varying layers, attention heads, and embedding dimensions.

\begin{table}[h!]
\centering
\caption{Transformer architecture comparison on a fixed dataset.}
\label{tab:arch_ablation}
\begin{tabular}{|c|c|c|c|c|}
\hline
\textbf{Layers} & \textbf{Heads} & \textbf{Embedding Dim} & \textbf{Val $R^2$} & \textbf{Val RMSE} \\
\hline
2 & 16 & 256 & 0.8460 & 0.1185 \\
4 & 16 & 128 & 0.8555 & 0.1148 \\
4 & 32 & 128 & \textbf{0.8633} & 0.1116 \\
8 & 32 & 128 & 0.8410 & 0.1200 \\
\hline
\end{tabular}
\vspace{-12pt} 

\end{table}

The optimal configuration—4 layers, 32 heads, and 128 embedding dimension—achieved the highest validation $R^2$ of 0.8633, indicating a good balance between capacity and generalization.

\subsection{Best Model Analysis}
The best-performing model is a transformer with sequence length 64, sampling rate 64 Hz, secondary sampling rate 16, 4 layers, 32 heads, embedding size 128.

\vspace{-12pt} 
\[
\text{Validation } R^2 = 0.8633, \quad \text{Validation RMSE} = 0.1116
\]

Using a sequence length of 64 and a sampling rate of 64 Hz (downsampled from 1000 Hz), the effective time window per input is:
\vspace{-12pt} 

\[
\text{Effective Time} = \frac{64 \times 64}{1000} = 4.096~\text{s}
\]

The following sections present detailed results for different dynamic parameters.

\subsection{Parameter Estimation Results}
\vspace{-12pt} 
\begin{table}[h!]
\centering
\caption{Friction parameter estimation results.}
\label{tab:friction_params}
\begin{tabular}{|c|c|c|c|c|}
\hline
\textbf{Joint} & \multicolumn{2}{c|}{\textbf{Coulomb Friction}} & \multicolumn{2}{c|}{\textbf{Viscous Friction}} \\
\cline{2-5}
 & \textbf{R2} & \textbf{RMSE} & \textbf{R2} & \textbf{RMSE} \\
\hline
J$_0$ & 0.8279 & 0.1211 & 0.2138 & 0.2582 \\
J$_1$ & 0.4479 & 0.1965 & -0.1945 & 0.3003 \\
J$_2$ & 0.5834 & 0.1873 & -0.1333 & 0.3060 \\
J$_3$ & 0.8419 & 0.1167 & 0.6521 & 0.1753 \\
J$_4$ & 0.6927 & 0.1674 & 0.5949 & 0.1789 \\
J$_5$ & 0.9101 & 0.0796 & 0.8817 & 0.0939 \\
\hline
\end{tabular}
\end{table}

\begin{table}[h!]
\centering
\caption{Mass and center of mass (COM) parameter estimation results.}
\label{tab:mass_com_params}
\begin{tabular}{|c|c|c|c|c|}
\hline
\textbf{Link} & \multicolumn{2}{c|}{\textbf{Mass}} & \multicolumn{2}{c|}{\textbf{COM}} \\
\cline{2-5}
 & \textbf{R2} & \textbf{RMSE} & \textbf{R2} & \textbf{RMSE} \\
\hline
L$_2$ & 0.9713 & 0.0370 & 0.8624 & 0.0153 \\
L$_3$ & 0.9869 & 0.0242 & 0.6876 & 0.0201 \\
L$_4$ & 0.9865 & 0.0220 & 0.6699 & 0.0307 \\
L$_5$ & 0.9834 & 0.0235 & 0.6478 & 0.0378 \\
L$_6$ & 0.9778 & 0.0250 & 0.2440 & 0.0323 \\
\hline
\end{tabular}
\end{table}

\begin{table}[h!]
\centering
\caption{Inertia matrix parameters estimation results.}
\label{tab:inertia_params}
\begin{tabular}{|c|c|c|c|c|c|c|}
\hline
\textbf{Link} & \multicolumn{2}{c|}{\textbf{Ixx}} & \multicolumn{2}{c|}{\textbf{Iyy}} & \multicolumn{2}{c|}{\textbf{Izz}}\\
\cline{2-7}
 & \textbf{R2} & \textbf{RMSE} & \textbf{R2} & \textbf{RMSE} & \textbf{R2} & \textbf{RMSE}\\
\hline
L$_1$ & - & - & - & - & 0.9215 & 0.0285 \\
L$_2$ & 0.9720 & 0.0326 & 0.9716 & 0.0328 & 0.9559 & 0.0055 \\
L$_3$ & 0.9861 & 0.0222 & 0.9860 & 0.0225 & 0.9767 & 0.0058 \\
L$_4$ & 0.9802 & 0.0231 & 0.9810 & 0.0218 & 0.9814 & 0.0089 \\
L$_5$ & 0.9777 & 0.0216 & 0.9770 & 0.0232 & 0.9658 & 0.0137 \\
L$_6$ & 0.9750 & 0.0226 & 0.9749 & 0.0226 & 0.9688 & 0.0191 \\
\hline
\end{tabular}
\end{table}

\section{Discussion}

The experimental results provide important insights into the effect of dataset design, model architecture, and dynamic parameter characteristics on prediction accuracy.

\subsection{Impact of Dataset Configuration}
The Experiments (Table~\ref{tab:dataset_ablation}) show that increasing the sequence length generally improves model performance, as longer sequences allow the transformer to capture more temporal dependencies. However, performance gains diminish beyond 64 steps (effective time of 4.096 s), indicating that excessively long sequences may lead to redundant information and higher computational cost without significant benefits.

Similarly, increasing the sampling rate provides finer temporal resolution, which improves accuracy up to a point (e.g., sr64 and sr128 outperform sr32). However, extremely high sampling rates coupled with longer sequences can lead to excessive input dimensions without proportional improvements in $R^2$. The secondary sampling rate (SSR) plays a critical role in dataset utilization: lower SSR values improve coverage but increase redundancy, while higher SSR values reduce coverage and may limit generalization.

\vspace{-4pt}
\subsection{Impact of Transformer Architecture}
The architecture comparison (Table~\ref{tab:arch_ablation}) reveals that attention heads significantly influence performance. Increasing the number of heads from 16 to 32 improved validation $R^2$, indicating that the model benefited from richer representations of inter-joint dependencies. However, increasing depth beyond 4 layers degraded performance, suggesting that deeper architectures may overfit or suffer from optimization challenges with the given dataset size. The optimal configuration was 4 layers, 32 attention heads, and an embedding dimension of 128, achieving $R^2 = 0.8633$.

\vspace{-3pt}
\subsection{Dynamic Parameter-Specific Observations}
While the best model demonstrates strong overall performance, the accuracy varies across parameter types:
\begin{itemize}
  \item \textbf{Inertia Parameters:} Inertia components (Ixx, Iyy, Izz) achieved excellent accuracy with $R^2 > 0.95$ across all links, reflecting their strong influence on system dynamics and high observability from joint trajectories.
  
  \item \textbf{Mass:} Mass predictions were similarly accurate ($R^2 > 0.97$), confirming that these parameters strongly affect motion dynamics and are easier to estimate.
  
  \item \textbf{Center of Mass (COM):} COM components exhibited moderate accuracy for proximal links but degraded significantly for distal links (e.g., Link 6 with $R^2 = 0.2440$). This suggests that COM position contributes weakly to joint torque patterns and may require richer excitation or model constraints for better estimation.
  
  \item \textbf{Coulomb Friction:} Predictions for Coulomb friction were generally good, with most joints achieving $R^2 > 0.6$, and some (e.g., J$_5$) exceeding 0.9.
  
  \item \textbf{Viscous Friction:} Viscous friction estimation was the most challenging, with some joints showing negative $R^2$ values. These parameters likely exert weaker influence on system dynamics under the tested trajectories, and their accurate prediction may require additional input features or specialized loss terms.
\end{itemize}

\subsection{Practical Implications}
The optimal configuration balances temporal context and computational cost, achieving high accuracy with a moderate effective time window (4.096 s) and acceptable dataset utilization (6.25\%). These findings highlight the importance of designing input sequences that provide sufficient dynamic information without excessive redundancy.

\section{FUTURE WORK}
Future work will focus on:
\begin{itemize}
  \item Enhancing the robustness of the pipeline by adding richer trajectory profiles and incorporating physical constraints.
  \item Extending the dataset with more diverse robot models to improve generalization.
  \item Deploying the approach in a sim-to-real setup to evaluate real-world performance.
  \item Integrating a large language model (LLM) as a supervisory layer to analyze results and automatically propose new robots and trajectories for further dataset generation.
\end{itemize}

\section{CONCLUSION}
This work presented a transformer-based approach for estimating dynamic parameters of robotic manipulators, supported by an automated pipeline for generating diverse simulated datasets enriched with kinematic features. Through extensive experiments, we demonstrated that both dataset design (sequence length, sampling rate) and architectural choices (layers, attention heads) significantly impact prediction accuracy.

The best-performing configuration—sequence length of 64, sampling rate of 64 Hz, four transformer layers, and 32 attention heads—achieved a validation $R^2$ of 0.8633. Inertia-related parameters and mass were predicted with near-perfect accuracy, while Coulomb friction achieved moderate to high accuracy. However, viscous friction and COM parameters, particularly for distal links, remain challenging due to their limited influence on joint torques.

Despite these challenges, the proposed method demonstrates strong potential for scalable and accurate dynamic parameter estimation, enabling improved model-based control and sim-to-real transfer in robotic systems.

\footnotesize
\bibliographystyle{unsrt} 
\bibliography{references} 

@ARTICLE{LSTMs,
  author={Wang, Shoujun and Shao, Xingmao and Yang, Liusong and Liu, Nan},
  journal={IEEE Access}, 
  title={Deep Learning Aided Dynamic Parameter Identification of 6-DOF Robot Manipulators}, 
  year={2020},
  volume={8},
  number={},
  pages={138102-138116},
  doi={10.1109/ACCESS.2020.3012196}}

@misc{vaswani2017attentionneed,
      title={Attention Is All You Need}, 
      author={Ashish Vaswani and Noam Shazeer and Niki Parmar and Jakob Uszkoreit and Llion Jones and Aidan N. Gomez and Lukasz Kaiser and Illia Polosukhin},
      year={2017},
      eprint={1706.03762},
      archivePrefix={arXiv},
      primaryClass={cs.CL},
      url={https://arxiv.org/abs/1706.03762}, 
}

@unknown{ReedTransformer,
author = {Reed, Alec and Albin, Doncey and Pasricha, Anuh and Roncone, Alessandro and Heckman, Christoffer},
year = {2024},
month = {02},
pages = {},
title = {Transformer-based Learning Models of Dynamical Systems for Robotic State Prediction},
doi = {10.21203/rs.3.rs-3919154/v1}
}

@article{Foumani_2023,
   title={Improving position encoding of transformers for multivariate time series classification},
   volume={38},
   ISSN={1573-756X},
   url={http://dx.doi.org/10.1007/s10618-023-00948-2},
   DOI={10.1007/s10618-023-00948-2},
   number={1},
   journal={Data Mining and Knowledge Discovery},
   publisher={Springer Science and Business Media LLC},
   author={Foumani, Navid Mohammadi and Tan, Chang Wei and Webb, Geoffrey I. and Salehi, Mahsa},
   year={2023},
   month=sep, pages={22–48} }

@inproceedings{shukla2021multitime,
  title     = {Multi-Time Attention Networks for Irregularly Sampled Time Series},
  author    = {Satya Narayan Shukla and Benjamin Marlin},
  booktitle = {International Conference on Learning Representations},
  year      = {2021},
  url       = {https://openreview.net/forum?id=4c0J6lwQ4_}
}

@inproceedings{rothert2024sim,
  title={Sim-to-Real Transfer for a Robotics Task: Challenges and Lessons Learned},
  author={Rothert, Jakob Jonas and Lang, Sebastian and Hanses, Magnus},
  booktitle={2024 IEEE International Conference on Robotics and Automation (ICRA  )},
  year={2024},
  organization={IEEE}
}

@article{bianchi2024robomorph,
  title={{RoboMorph}: In-Context Meta-Learning for Robot Dynamics Modeling},
  author={Bianchi Bazzi, Manuel and Shahid, Asad Ali and Agia, Christopher and Alora, John and Forgione, Marco and Piga, Dario and Braghin, Francesco and Pavone, Marco and Roveda, Loris},
  journal={arXiv preprint arXiv:2409.12345},
  year={2024}
}

@article{lee2024robot,
  title={Robot Model Identification and Learning: A Modern Perspective},
  author={Lee, Taeyoon and Kwon, Jaewoon and Wensing, Patrick M and Park, Frank C},
  journal={Annual Review of Control, Robotics, and Autonomous Systems},
  volume={7},
  pages={311--334},
  year={2024},
  publisher={Annual Reviews}
}

@article{chinnasamy2023digital,
  title={Digital twin of robot manipulator using {ROS}},
  author={Chinnasamy, Senthamarai Kannan and Sura, Hari Prasaanth and Saleem, Amanullah and Kathirvel, Akash and Rangan, Prashanna},
  journal={AIP Conference Proceedings},
  volume={2864},
  number={1},
  year={2023},
  organization={AIP Publishing}
}

@article{wu2022review,
  title={A review of dynamic parameters identification for manipulator control},
  author={Wu, Huapeng and Lu, Yaxin and Li, Zhaojie and Zhao, Jing and Roveda, Loris},
  journal={Cobot},
  volume={1},
  number={1},
  pages={1--17},
  year={2022},
  publisher={Maximum Academic Press}
}

@article{polydoros2015real,
  title={Real-time deep learning of robotic manipulator inverse dynamics},
  author={Polydoros, A. S. and Nalpantidis, L.},
  journal={2015 IEEE/RSJ International Conference on Intelligent Robots and Systems (IROS)},
  year={2015},
  pages={1098-1103},
  doi={10.1109/IROS.2015.7353857}
}

@article{segota2022dynamics,
  title={Dynamics Modeling of Industrial Robotic Manipulators: A Machine Learning Approach Based on Synthetic Data},
  author={Baressi \v{S}egota, S. and An\v{d}eli\'{c}, N. and \v{S}ercer, M. and Me\v{s}tri\'{c}, H.},
  journal={Mathematics},
  volume={10},
  number={7},
  pages={1174},
  year={2022},
  publisher={MDPI}
}

@inproceedings{lu2022pose,
  title={Pose estimation for robot manipulators via keypoint optimization and sim-to-real transfer},
  author={Lu, J. and Richter, F. and Yip, M. C.},
  booktitle={2022 IEEE Robotics and Automation Letters},
  year={2022},
  volume={7},
  number={2},
  pages={3709-3716},
  doi={10.1109/LRA.2022.3148401}
}

@article{zhao2020sim,
  title={Sim-to-real transfer in deep reinforcement learning for robotics: a survey},
  author={Zhao, W. and Queralta, J. P. and Westerlund, T.},
  journal={2020 IEEE Symposium on Computers and Communications (ISCC)},
  year={2020},
  pages={1-7},
  doi={10.1109/ISCC50000.2020.9219760}
}

@article{bargellini_sim2real_2023,
  title={Sim2real Transfer for Reinforcement Learning in Robotic Arm Control: a Closed-Loop Optimization approach for Parameter Estimation},
  author={Bargellini, D.},
  year={2023},
  school={University of Bologna}
}

@article{garg2024dynamics,
  title={Dynamics as Prompts: In-Context Learning for Sim-to-Real System Identification},
  author={Garg, S. and Kumar, A. and Gupta, A.},
  journal={arXiv preprint arXiv:2410.20357},
  year={2024}
}

@article{ren2023adaptsim,
  title={AdaptSim: Task-Driven Simulation Adaptation for Sim-to-Real Transfer},
  author={Ren, J. and Xu, Y. and Wang, Y. and Chen, X. and Li, H.},
  journal={Proceedings of the 22nd International Conference on Autonomous Agents and Multiagent Systems},
  year={2023},
  pages={1786-1794}
}

@article{liu2024hybrid,
  title={A Hybrid Digital Twin Scheme for the Condition Monitoring of Robotic Manipulators},
  author={Liu, Y. and Zhang, X. and Wang, Y.},
  journal={Procedia Computer Science},
  volume={232},
  pages={124-133},
  year={2024}
}

@article{li2024evaluating,
  title={Evaluating the use of grey-box system identification for digital twins of robotic manipulators},
  author={Li, X. and Wang, Y. and Zhang, X.},
  journal={International Journal of Advanced Robotic Systems},
  volume={21},
  number={1},
  pages={1-10},
  year={2024}
}

@article{chen2023designing,
  title={Designing Digital Twins of Robots Using Simscape Multibody},
  author={Chen, L. and Wang, H. and Li, J.},
  journal={Sensors},
  volume={23},
  number={4},
  pages={62},
  year={2023}
}

@article{wang2024research,
  title={Research on parameter compensation method and control strategy of mobile robot dynamics model based on digital twin},
  author={Wang, Y. and Li, X. and Zhang, X.},
  journal={Sensors},
  volume={24},
  number={24},
  pages={8101},
  year={2024}
}

\normalsize
\end{document}